\documentclass[10pt,twocolumn,letterpaper]{article}

\usepackage{iccv}
\usepackage{times}
\usepackage{epsfig}
\usepackage{graphicx}
\usepackage{amsmath}
\usepackage{amssymb}

\newcommand{\ourmethod}{ImGeoNet}

\newcommand{\mdfootnotesize}[1]{\begin{footnotesize}#1\end{footnotesize}}

\newcommand{\mdvec}[1]{\mathbold{#1}}
\newcommand{\mdmat}[1]{\mathbold{#1}}

\newcommand{\mdineq}[1]{\(#1\)}
\newcommand{\E}{\mathrm{E}}
\newcommand{\Var}{\mathrm{Var}}

\newcommand{\mdsec}[1]{Sec.~#1}
\newcommand{\mdfig}[1]{Fig.~#1}
\newcommand{\mdeq}[1]{Eq.~#1}
\newcommand{\mdtbl}[1]{Table.~#1}
\newcommand{\mdapp}[1]{Appendix.~#1}
\newcommand{\mdsubsec}[1]{\vspace{3pt}\noindent\textbf{#1}}

\usepackage{fixmath}
\usepackage{booktabs}
\usepackage{enumitem}

\usepackage[pagebackref=true,breaklinks=true,letterpaper=true,colorlinks,bookmarks=false]{hyperref}

\iccvfinalcopy 
\def\iccvPaperID{2631} 

\ificcvfinal\fi

\begin{document}

\title{
{\ourmethod}: Image-induced Geometry-aware Voxel Representation for Multi-view 3D Object Detection}

\author{
Tao Tu$^1$ \quad Shun-Po Chuang$^2$ \quad Yu-Lun Liu$^3$ \quad Cheng Sun$^1$ \quad Ke Zhang$^4$ \\ Donna Roy$^{4}$\thanks{Work done in Amazon.} \quad Cheng-Hao Kuo$^4$ \quad Min Sun$^{1,4}$
\vspace{15pt} \\
$^1$National Tsing Hua University \quad $^2$National Taiwan University \\ $^3$National Yang Ming Chiao Tung University \quad $^4$Amazon \vspace{15pt} \\
{\tt\small ttaoretw@gmail.com \quad f04942141@ntu.edu.tw \quad yulunliu@cs.nycu.edu.tw}\\
{\tt\small chengsun@gapp.nthu.edu.tw \quad kezha@amazon.com \quad donna.v.roy@gmail.com}\\
{\tt\small chkuo@amazon.com \quad sunmin@ee.nthu.edu.tw}
}

\maketitle
\ificcvfinal\thispagestyle{empty}\fi
\begin{abstract}
  We propose {\ourmethod}, a multi-view image-based 3D object detection framework that models a 3D space by an image-induced geometry-aware voxel representation.
  Unlike previous methods which aggregate 2D features into 3D voxels without considering geometry, {\ourmethod} learns to induce geometry from multi-view images to alleviate the confusion arising from voxels of free space, and during the inference phase, only images from multiple views are required.
  Besides, a powerful pre-trained 2D feature extractor can be leveraged by our representation, leading to a more robust performance.
  To evaluate the effectiveness of {\ourmethod}, we conduct quantitative and qualitative experiments on three indoor datasets, namely ARKitScenes, ScanNetV2, and ScanNet200.
  The results demonstrate that {\ourmethod} outperforms the current state-of-the-art multi-view image-based method, ImVoxelNet, on all three datasets in terms of detection accuracy.
  In addition, {\ourmethod} shows great data efficiency by achieving results comparable to ImVoxelNet with 100 views while utilizing only 40 views.
  Furthermore, our studies indicate that our proposed image-induced geometry-aware representation can enable image-based methods to attain superior detection accuracy than the seminal point cloud-based method, VoteNet, in two practical scenarios: (1) scenarios where point clouds are sparse and noisy, such as in ARKitScenes, and (2) scenarios involve diverse object classes, particularly classes of small objects, as in the case in ScanNet200.
  Project page: \url{https://ttaoretw.github.io/imgeonet}.

\end{abstract}

\section{Introduction}
  \begin{figure}[t]
\begin{center}
  \includegraphics[width=\linewidth]{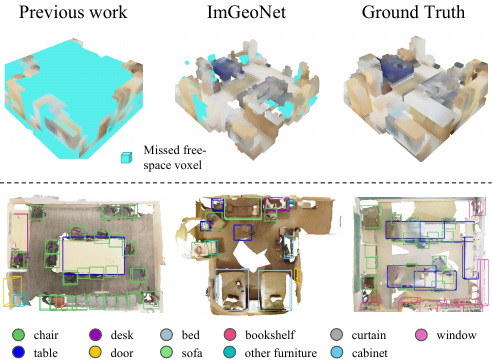}
\end{center}
\vspace{-2pt}
  \caption{\textbf{Geometry-aware voxel representation.}
  (\emph{Top part}) In contrast to prior works~\cite{rukhovich2022imvoxelnet} (\emph{top left}) that disregard the underlying geometry, our proposed {\ourmethod} (\emph{top center}) successfully preserves the geometric structure with respect to the ground truth (\emph{top right}) while effectively reducing the number of voxels in free space.
  In the visualization of {\ourmethod}, voxels with a surface probability exceeding a predefined threshold are retained, otherwise removed. The color of each voxel is determined by averaging the colors of ground truth point clouds within the voxel.
  Missed free-space voxels are marked as cyan.
  (\emph{Bottom part}) We present the \textbf{detection results} using bounding cubes that are color-coded based on the predicted categories.
  }
\label{fig:teaser}
\end{figure}

  Indoor 3D object detection has been an active area of computer vision research for over a decade, owing to its practical applications in robotics, augmented reality, and mixed reality.
  In recent years, several studies~\cite{qi2019votenet, hou20193dsis, cheng2021brnet, zhang2020h3dnet, wang2022rbgnet, liu2021groupfree, misra2021-3detr, gwak2020gsdn, rukhovich2022fcaf3d} have demonstrated the effectiveness of methods based on point clouds in conjunction with deep learning techniques for indoor 3D object detection.
  However, the applicability of these methods is limited by their reliance on data acquired from expensive 3D sensors such as depth cameras, stereo cameras, or laser scanners.
  In contrast to point clouds, color images are more affordable and can capture semantically rich information akin to human vision.
  Therefore, image-based indoor 3D object detection is a promising research direction.

  Image-based methods for indoor monocular 3D object detection~\cite{huang2019perspectivenet, huang2018cooperative, nie2020total3dunderstanding, zhang2021im3d} have demonstrated a satisfactory level of accuracy.
  Nonetheless, monocular methods encounter challenges such as scale ambiguity, occlusion issues, and limited field of view.
  These issues can be mitigated by providing multiple perspectives of the scene, leading to a more robust and accurate 3D object detection result.
  Previous works~\cite{murez2020atlas, rukhovich2022imvoxelnet} employ multi-view images to construct a feature volume, which is subsequently utilized for conducting 3D object detection~\cite{rukhovich2022imvoxelnet}.
  Although these methods have exhibited state-of-the-art performance, they neglect the underlying geometric characteristics during the feature volume construction.
  
  In this work, we propose {\ourmethod}, a multi-view 3D object detection framework that models a 3D space by an image-induced geometry-aware voxel representation.
  {\ourmethod} learns to induce geometry from multi-view images to reduce the importance of voxels representing free space, and during the inference phase, only images from multiple views are required.
  Specifically, {\ourmethod} predicts the likelihood of each voxel belonging to a surface, and subsequently weighting the feature volume according to this probability.
  The proposed approach exhibits a notable enhancement in detection performance owing to the successful alleviation of confusion arising from voxels in free space.
  Besides, a powerful pre-trained 2D feature extractor can be utilized by our representation, leading to more robust performance.

  We conduct quantitative and qualitative experiments to evaluate the effectiveness of {\ourmethod} on three indoor datasets, namely ARKitScenes~\cite{baruch2021arkitscenes}, ScanNetV2~\cite{dai2017scannet}, and ScanNet200~\cite{rozenberszki2022scannet200}.
  The results demonstrate that {\ourmethod} outperforms the state-of-the-art multi-view image-based method, ImVoxelNet~\cite{rukhovich2022imvoxelnet}, by $3.8\%$, $12.5\%$ and $17.4\%$ in mAP@0.25 on ARKitScenes, ScanNetV2, and ScanNet200, respectively.
  Additionally, {\ourmethod} shows great data efficiency by achieving results comparable to ImVoxelNet with 100 views while utilizing only 40 input views.
  Furthermore, the results of the experiments indicate that our proposed image-induced geometry-aware representation can enable image-based methods to attain superior detection accuracy than the seminal point cloud-based method, VoteNet, in two practical scenarios: (1) scenarios where point clouds are sparse and noisy, such as in ARKitScenes, and (2) scenarios involve diverse object classes, particularly classes of small objects, as in the case in ScanNet200.
  Specifically, {\ourmethod} outperforms VoteNet in these scenarios by at least $12.6\%$ in terms of mAP@0.25.
  
  The contributions of our work can be summarized as follows:
  \begin{itemize}[itemsep=1pt, topsep=0pt, parsep=0pt]
    \item We introduce a multi-view object detection framework that utilizes an image-induced geometry-aware voxel representation to enhance image-based 3D object detection substantially.
    \item Our method achieves state-of-the-art performance for image-based 3D object detection on ARKitScenes, ScanNetV2, and ScanNet200.
    \item Our studies demonstrate our proposed geometry-aware representation enables image-based methods to attain superior detection accuracy than the seminal point cloud-based method, VoteNet, in practical scenarios which consist of sparse and noisy point clouds or involve diverse classes.
  \end{itemize}

\begin{figure*}[ht!]
\begin{center}
  \includegraphics[width=\textwidth]{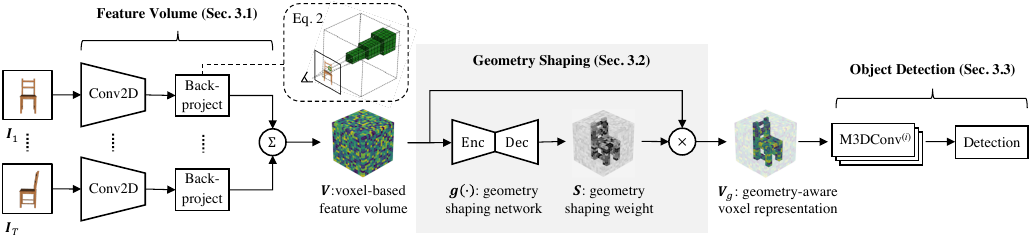}
\end{center}
\vspace{-2pt}
\caption{
  \textbf{An illustration of {\ourmethod} framework} for 3D object detection.
  Given an arbitrary number of images, a 2D convolution backbone (Conv2D) is applied to extract visual features from each image, and then a 3D voxel feature volume (\mdineq{\mdmat{V}}) is constructed by back-projecting (Eq.~\ref{eq:backproj-vol-copy}) and accumulating 2D features to the volume (\mdsec{\ref{approach:vx-feat-vol}}).
  This feature volume is not ideal since the underlying geometry of the scene is not considered.
  Hence, geometry shaping (\mdsec{\ref{approach:geo-shaping}}) is applied to weight the original feature volume by the predicted surface probabilities (\mdineq{\mdmat{S}}), which preserves the geometric structure and removes voxels of free space.
  Finally, the geometry-aware volume (\mdineq{\mdmat{V}_g}) is passed to the multiscale 3D convolutional layers (M3DConv) and the detection head (\mdsec{\ref{approach:obj-det}}).
}
\label{fig:method}
\end{figure*}

\section{Related Work}  
  \subsection{Point Cloud Based Object Detection}
  Since point clouds provide reliable geometric structure information, point cloud-based object detection has shown great performance on both indoor and outdoor scenes.
  There are two main branches, one is point-based methods directly sampling based on the set abstract and the feature propagation~\cite{qi2017pointnet++, qi2018fpointnet, yang2019std, shi2019pointrcnn, yang20203dssd, shi2020pointgnn, pan2021pointformer, qi2019votenet, zhang2020h3dnet} while the other is grid-based methods based on grid representation~\cite{yang2018pixor, zhou2018voxelnet, yan2018second, lang2019pointpillars, shi2020pillarod, deng2021voxelrcnn, mao2021voxeltransformer, gwak2020gsdn, rukhovich2022fcaf3d}.
  
  As far as indoor object detection is concerned, the predominant methods~\cite{qi2019votenet, cheng2021brnet, zhang2020h3dnet, wang2022rbgnet} are those relying on deep Hough voting~\cite{qi2019votenet} to shift surface points to their corresponding object centers.
  Transformer-based methods~\cite{misra2021-3detr, liu2021groupfree} also deliver comparable results.
  Recently, sparse fully-convolutional detection methods~\cite{gwak2020gsdn, rukhovich2022fcaf3d} have exhibited state-of-the-art performance with regard to both accuracy and efficiency.
  Despite the fact that point cloud-based methods perform well on object detection, they rely on costly 3D sensors, which narrows down their use cases.

  \subsection{Image-based Object Detection}
\mdsubsec{Monocular Object Detection.}
  There has been considerable attention paid to the field of monocular object detection due to its practicality and cost-effectiveness.
  Two-stage methods~\cite{simonelli2019disentangling, shi2021geometry, qin2019monogrnet, manhardt2019roi, lu2021geometry, li2019gs3d} extend conventional two-stage 2D detection frameworks to estimate 3D object parameters.
  Single-stage anchor-based methods~\cite{brazil2019m3d, brazil2020kinematic, kumar2021groomed, liu2019deep, luo2021m3dssd} and anchor-free methods~\cite{li2020rtm3d, liu2020smoke, ma2021delving, mousavian20173d, roddick2018orthographic, wang2021fcos3d, wang2022probabilistic, zhang2021objects, zhou2019objects, zhou2021monocular} predict object parameters in one stage.
  To mitigate the depth information loss, some approaches ~\cite{reading2021cadnn, ding2020learning, park2021pseudo, wang2021depth, xu2018multi} use an additional backbone for depth map feature extraction, and other approaches~\cite{ma2019accurate, wang2020task, wang2019pseudo} back-project depth images to 3D pseudo point clouds.
  The improvements highlight the significance of depth information in 3D object detection.
  
  As for the indoor environment, some prior works~\cite{zhao2011image, zhao2013scene, choi2013understanding, lin2013holistic, zhang2014panocontext} estimate the 3D bounding boxes based on geometry and 3D world priors.
  Other works~\cite{huang2018holistic, izadinia2017im2cad} utilize the category-specific 3D shape for detecting objects.
  More recently, several works~\cite{song2015sunrgbd, huang2018cooperative, nie2020total3dunderstanding, zhang2021im3d} view object detection as a component of scene understanding.
  However, they face challenges associated with monocular images, including scale ambiguity, occlusion problems and limited field of view.
  
\mdsubsec{Multi-view Object Detection.}
  To better capture scene information, methods that consider multiple views has gained increasing attention in recent years.
  DETR-based methods~\cite{wang2022detr3d, liu2022petr, tseng2022crossdtr} extend DETR~\cite{carion2020detr} to 3D object detection.
  Besides, prior research~\cite{huang2021bevdet, li2022bevformer} has demonstrated that the bird-eye-view (BEV) representation is well-suited for object detection in autonomous driving scenarios.
  Although the aforementioned methods perform well in autonomous driving scenarios, they may not be applicable to indoor scenes, which often contain diverse object classes that are not necessarily situated on the ground.
  ImVoxelNet~\cite{rukhovich2022imvoxelnet}, on the other hand, has shown great performance in the domain of indoor 3D object detection by performing on a 3D voxel-based feature volume~\cite{murez2020atlas}.
  However, it does not properly preserve the underlying geometry of input scenes during the feature volume construction.

\section{Approach}
\label{approach}
  In this section, we first introduce the problem formulation, provide an overview of our method {\ourmethod} and briefly explain the design concept.
  Next, we present the main steps of {\ourmethod} (\mdfig{\ref{fig:method}}) in detail, which include feature volume construction (\mdsec{\ref{approach:vx-feat-vol}}), geometry shaping (\mdsec{\ref{approach:geo-shaping}}) and object detection (\mdsec{\ref{approach:obj-det}}).

  \mdsubsec{Problem Formulation.}
  Given an arbitrary number of input images captured in the same scene \mdineq{\{\mdmat{I}_t\} \subseteq {\mathbb{R}}^{H \times W \times 3}} and their corresponding intrinsic matrices \mdineq{\{\mdmat{K}_t\} \subseteq {\mathbb{R}}^{3 \times 3}} and poses \mdineq{\{\mdmat{T}_t\} \subseteq \mathrm{SE}(3)}, the goal of 3D object detection is to identify target objects by predicted categories and enclosing bounding boxes \mdineq{\{\mdvec{b}\} \subseteq {\mathbb{R}}^{7}}.
  A bounding box is parameterized by \mdineq{(x, y, z, w, h, l, {\phi})}, where \mdineq{(x, y, z)} is the center, \mdineq{(w, h, l)} is the size and \mdineq{\phi} is the yaw angle.
  We follow a common assumption~\cite{qi2019votenet, rukhovich2022imvoxelnet} that bounding boxes are on the ground plane, so only yaw angles are predicted.
  
  \mdsubsec{Framework Overview.}
  {\ourmethod} aims to predict the 3D bounding boxes and corresponding object categories for the target objects present in the scene, using an arbitrary number of input images.
  
  First of all, {\ourmethod} constructs a 3D voxel feature volume by back-projecting and accumulating 2D features to the volume as in \cite{murez2020atlas, rukhovich2022imvoxelnet}.
  The 2D feature of a pixel is duplicated to voxels along the ray emitted from the camera center through the pixel.
  The 3D voxel volume obtained in this process is suboptimal, as it can lead to contamination of voxels in free space and hinder the precision of detection.
  
  To turn the constructed voxel volume into a geometry-aware representation, {\ourmethod} performs geometry shaping.
  In this step, {\ourmethod} weights each voxel feature according to the probability of that voxel being located on an object's surface.
  Consequently, voxels situated in free space will be assigned a lower weight, whereas those located on object surfaces will retain a higher weight and incorporate information from various viewpoints.
  This significantly improves the accuracy of the predicted bounding box.
  During training, we convert the ground-truth point clouds to surface voxels to supervise geometry shaping network.
  Finally, we follow previous works~\cite{rukhovich2022imvoxelnet, tian2020fcos} to predict the bounding box for each voxel and perform non-maximal suppression to reduce redundant predictions.

\subsection{Feature Volume}
\label{approach:vx-feat-vol}
  Attaining precise 3D object detection requires a thorough comprehension of the geometric structure inherent to a given scene.
  The feature volume approaches have been demonstrated to be highly effective in tasks that necessitate an extensive understanding of scene geometry, such as stereo matching~\cite{im2019dpsnet, yao2018mvsnet}, surface reconstruction~\cite{dai2018scancomplete, muller2022instant, yu2022monosdf}, and novel view synthesis~\cite{chen2021mvsnerf, sun2022direct}.
  Consequently, we adopt the feature volume representation to describe a scene, and leverage a pre-trained 2D feature extractor for more robust performance.

  We compute the feature volume \mdineq{\mdmat{V} \in \mathbb{R}^{H_v \times W_v \times D_v \times C}} from a sequence of images \mdineq{\mdmat{I}_t} with known camera intrinsics \mdineq{\mdmat{K}_t} and poses \mdineq{\mdmat{T}_t}.
  Here \mdineq{H_v}, \mdineq{W_v} and \mdineq{D_v} denote the side lengths of the volume in terms of the voxel size unit, while \mdineq{C} represents the feature dimension.
  We first extract 2D features by 2D convolutional backbone from input images
  \begin{equation}
  \label{eq:2d-feat-extraction}
    \mdmat{F}_t = \text{Backbone2D}(\mdmat{I}_t),
  \end{equation}
  where \mdineq{\mdmat{F}_t \in \mathbb{R}^{H \times W \times C}} and strides in convolution layers are ignored for simplicity.
  Next, the 2D features are back-projected to the volume by
  \begin{equation}
  \label{eq:backproj-vol-copy}
    \mdmat{V}_t[x, y, z,:] = \mdmat{F}_t[u, v, :],
  \end{equation}
  where \mdineq{[:]} is the slice operator and the pixel coordinates \mdineq{(u, v)} are computed from the voxel centers \mdineq{(x, y, z)} by the pinhole camera model as follows:
  \begin{equation}
  \label{eq:backproj-coord}
    \begin{bmatrix}
    u \\
    v \\
    1
    \end{bmatrix}
    = \frac{1}{\lambda} \mdmat{K}_t \mdmat{\Pi}_0 \mdmat{T}_t 
    \begin{bmatrix}
    x \\
    y \\
    z \\
    1
    \end{bmatrix},
    \text{~~where~}
    \mdmat{\Pi}_0 =
    \begin{bmatrix}
    1&0&0&0 \\
    0&1&0&0 \\
    0&0&1&0
    \end{bmatrix},
  \end{equation}
  and \mdineq{\lambda} is the distance along optical axis between the voxel center and the camera center.
  In practice, we scan over the voxel centers and retrieve the corresponding back-projected 2D features.
  If any voxels happen to be situated outside the view frustum, their features are assigned a value of zero.
  Finally, the back-projected volumes \mdineq{\mdmat{V}_{t}} computed from different views in \mdeq{\ref{eq:backproj-vol-copy}} are averaged to construct the feature volume \mdineq{\mdmat{V}} by
  \begin{equation}
  \label{eq:vol-merge}
  \mdmat{V} = (\frac{1}{\sum_{t} \mdmat{M}_t}) \odot (\sum_{t} \mdmat{V}_t \odot \mdmat{M}_t),
  \end{equation}
  where \mdineq{\odot} is the Hadamard product, and \mdineq{\mdmat{M}_t} is a binary mask indicating whether the voxels are in the view frustum of \mdineq{\mdmat{I}_t}.
  
  It is worth noting that the feature volume obtained through the mentioned construction process lacks information on the geometric structure of the scene.
  Specifically, each voxel along a camera ray is assigned the same feature value as its corresponding pixel, regardless of whether the voxel is located on the closest surface along that ray.
  Therefore, to properly incorporate the scene geometry, we introduce \emph{Geometry Shaping}.

\subsection{Geometry Shaping}
\label{approach:geo-shaping}
  One significant shortcoming of the derived feature volume from \mdsec{\ref{approach:vx-feat-vol}} is its geometry-unaware nature.
  In other words, even the voxels that are not on any object surface are still assigned values.
  The situation worsens when the voxels are in free space, where the detection module can be perturbed and generate false predictions.
  To address this, we propose \textit{geometry shaping}, which leverages the multi-view image input to induce geometry structure and remove noisy voxels in free space by down-weighting unoccupied voxels while preserving voxels on surfaces.

  Since appearance variance reveals certain geometric information~\cite{yao2018mvsnet, chen2021mvsnerf}, we also take the feature variance into consideration.
  To be specific, we concatenate feature variance (refer to \mdeq{\ref{eq:vol-merge-var}} in \mdsec{\ref{sec:impl}}) with the feature volume to obtain \mdineq{\mdmat{V}^{'}}.
  Subsequently, a geometry shaping volume \mdineq{\mdmat{S}} is generated via the geometry shaping network \mdineq{g(\cdot)} by
  \begin{equation}
  \label{eq:geo-shaping-grid}
    \mdmat{S} = g(\mdmat{V}^{'}),
  \end{equation}
  where \mdineq{\mdmat{S}} shares the same grid size as the combined feature volume \mdineq{\mdmat{V}^{'}}, and each element of \mdineq{\mdmat{S}} is the likelihood of the voxel being on an object surface.
  Note that $g(\cdot)$ employs the same feature volume (prior to geometry shaping) as the final 3D object detector, resulting in less overhead compared to computing Multi-View Stereo (MVS) from raw images.
  Afterward, the geometry-aware feature volume is derived by weighting the feature volume as follows:
  \begin{equation}
  \label{eq:geo-shaping}
    \mdmat{V}_{g} = \mdmat{S} \odot \mdmat{V}.
  \end{equation}
  Since the weights will be low in free space, the resulting geometry-aware feature volume mainly remains values on object surfaces, which could better describe the underlying geometry.
  As a result, the geometry shaping reduces the burden of the final detection to a great extent and improves precision.
  
  In our implementation, we convert the RGB-D frames to point clouds and consider the voxels that contain at least one point as surface voxels.
  Besides, for each camera ray, we also consider locations neighboring surface voxels within a margin of \mdineq{\epsilon} as positive.
  Subsequently, we proceed to supervise the geometry shaping network through surface voxel prediction using focal loss~\cite{lin2017focal} in an end-to-end way.
  It is worth noting that depth sensory data is solely employed to supervise the geometry shaping network during the training phase, while only images from multiple views are utilized during the inference phase.

\subsection{Object Detection}
\label{approach:obj-det}
  Even though the obtained feature volume in \mdeq{\ref{eq:geo-shaping}} is geometry-aware, it may still have limitations in capturing objects of varying scales.
  Therefore, we transform the geometry-aware volume by multiscale dense 3D convolution layers:
  \begin{equation}
  \label{eq:neck}
    \mdmat{V}_h^{(i)} = \text{M3DConv}^{(i)}(\mdmat{V}_g),
  \end{equation}
  where \mdineq{i \in \{0,1,...,L-1\}} is the scale index, and the volume of different scales will have different grid sizes.

  As regards the detection head, we follow ImVoxelNet~\cite{rukhovich2022imvoxelnet} to extend the single-stage anchor-free 2D detectors~\cite{tian2020fcos, zhang2020bridging} to 3D volume.
  All locations from \mdineq{L} scales are considered, and for each location, a class probability, center-ness, and a 3D bounding box are predicted.
  However, only a few locations are selected as positive samples for supervision during training:
  1) Locations not in any target bounding box are removed.
  2) For each target object, only locations from the most fitting scale are kept. Specifically, we choose the smallest scale that contains more than \mdineq{M} points.
  3) For each target object, we only keep the top-\mdineq{k} locations close to the bounding box center.
  4) If a point corresponds to multiple targets, we choose the one with minimal volume.
  Finally, we use focal loss~\cite{lin2017focal} for category classification, cross-entropy loss for center-ness estimation~\cite{tian2020fcos}, and rotated 3D IoU loss~\cite{zhou2019iou} for bounding box prediction.

\section{Implementation}
\label{sec:impl}
\begin{figure}[h!]
\begin{center}
  \includegraphics[width=\linewidth]{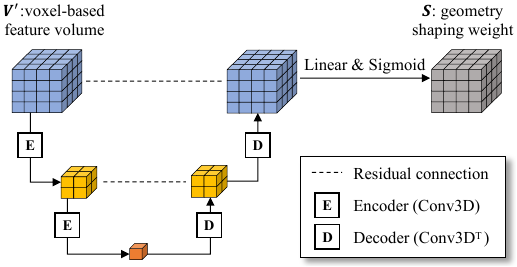}
\end{center}

\vspace{-2pt}

\caption{\textbf{The network architecture of geometry shaping network}. It is based on an encoder-decoder architecture with residual connections, followed by a linear projection layer and a sigmoid function.
}
\label{appendix:fig:network}
\end{figure}

  \mdsubsec{Geometry Shaping Network.}
  As illustrated in \mdfig{\ref{appendix:fig:network}}, our geometry shaping network has an encoder-decoder architecture with residual connections.
  Specifically, each encoder comprises of three 3D convolutional layers, while each decoder comprises of one transposed 3D convolutional layer and one 3D convolutional layer.
  We set the kernel sizes of all the 3D convolutional layers to 3, and set the kernel size of the transposed 3D convolutional layer to 2.
  Following each convolutional and transposed convolutional layer, we apply a batch normalization and ReLU activation.
  During encoding, the spatial sizes are reduced by a factor of 2, while the channel size is increased by a factor of 2.
  In contrast, during decoding, the spatial sizes are increased, and the channel size is reduced.
  Finally, we adopt a linear projection layer to reduce the channel size to 1, and we obtain the final output by passing the reduced tensor to a sigmoid function.

  \mdsubsec{Framework Architecture.}
  A ResNet50~\cite{he2016resnet} pre-trained on ImageNet~\cite{russakovsky2015imagenet} is used as the 2D feature extractor.
  The surface voxel margin \mdineq{\epsilon} is set to 4 voxels. 
  The M3DConv network is the same as the 3D convolutional network before the detection head in ImVoxelNet~\cite{rukhovich2022imvoxelnet}.
  As for the detection head, we follow the same previous work~\cite{rukhovich2022imvoxelnet} and set the number of scales \mdineq{L} to 3, the minimum fitting points \mdineq{M} to 27 and \mdineq{k} in the top-\mdineq{k} selection to 18.
  It is worth mentioning that {\ourmethod} is trained in an end-to-end way.
  Specifically, the geometry shaping network and all the remaining networks are jointly optimized for both surface voxel prediction and 3D object detection.
  

  \mdsubsec{Feature Variance.}
  We compute the feature variance across different views for each voxel as follows:
  \begin{equation}
  \small
    \label{eq:vol-merge-var}
    \mdmat{V}_\text{var} = \left(\frac{1}{\sum_{t} \mdmat{M}_t} \odot \left(\sum_{t} (\mdmat{V}_t)^2 \odot \mdmat{M}_t\right)\right) - (\mdmat{V}_\text{mean})^2,
  \end{equation}
  where {\small \mdineq{\mdmat{V}_\text{mean}}} is computed via \mdeq{\ref{eq:vol-merge}} in the paper and we utilize the fact that \mdineq{\Var(X)=\E[X^2]-\E[X]^2}.
  Subsequently, our geometry shaping network utilizes the concatenated values of \mdineq{\mdmat{V}_\text{mean}} and \mdineq{\mdmat{V}_\text{var}} as input for predicting surface voxels.
  
  \mdsubsec{Loss Configuration.}
  Both the surface voxel prediction loss and category classification loss are focal losses with a gamma value of 2 and an alpha value of 0.25.
  The centerness estimation loss is a cross-entropy loss, and the bounding box prediction loss is a rotated 3D IoU loss. All the loss weights are set to 1, except for the surface voxel prediction loss, which is assigned a weight of 10.
  It is worth mentioning that we employed identical loss hyperparameters across all experiments.

  

\section{Experiment}
\label{sec:exp}
  We describe the datasets, evaluation metric, and implementation details in Sec.~\ref{subsec:setup}.
  Then, we present and analyze the results of our experiments in Sec.~\ref{subsec:results}.

  \subsection{Setup}
\label{subsec:setup}
  \mdsubsec{Dataset.} Our approach is evaluated on three indoor multi-view datasets, each serving a distinct purpose.
  Firstly, to examine the feasibility of the proposed method in a realistic scenario, we use ARKitScenes~\cite{baruch2021arkitscenes} which contains sensor data captured by popular mobile devices and is the most realistic of the three datasets.
  Secondly, ScanNetV2~\cite{dai2017scannet} is used as it is the most widely adopted benchmark for comparing against other state-of-the-art methods.
  Lastly, to investigate the model performance in a setting with diverse classes of varying sizes and properties, ScanNet200~\cite{rozenberszki2022scannet200} is adopted since it comprises the largest number of object categories.

  \textbf{ARKitScenes}~\cite{baruch2021arkitscenes} is a large real-world RGB-D video dataset captured with handheld 2020 Apple iPad Pros.
  It comprises 5,047 captures of 1,661 distinct scenes, and we follow the official split to separate it into 4,498 and 549 captures for training and testing, respectively.
  The image resolution is 192\mdineq{\times}256.
  We uniformly sample the views for each scene based on the frame indices.
  The number of sampled views for training is 50 and 200 for the image-based methods and point cloud-based methods, respectively.
  To acquire the point clouds, we back-project the sampled views to 3D space based on the supplied low-resolution depth maps.
  We filter the points through voxel downsampling where the voxel size is set to 0.02 meters.
  Finally, it is noteworthy that the quality of point clouds in ARKitScenes is inferior to that of ScanNetV2 since the depth maps in ARKitScenes are low-resolution while the point clouds in ScanNetV2 are high-resolution and derived from the 3D reconstructed meshes.

  \textbf{ScanNetV2}~\cite{dai2017scannet} is a richly annotated RGB-D video dataset of 3D reconstructed meshes of indoor scenes.
  It contains 2.5 million views in 1,513 room-level scenes and there are 18 classes available for classification.
  The resolution of the images is 968\mdineq{\times}1296.
  We follow the public train-test split originally proposed by ScanNet, which allocates 1,201 scenes for training and 312 scenes for testing.
  Since the standard release does not provide oriented bounding box annotation, we follow \cite{hou20193dsis, qi2019votenet} to create axis-aligned bounding boxes according to the semantic labels of mesh vertices.
  To train and evaluate the image-based methods, we evenly sample 50 views for each scene based on the frame indices and resize the images to 480 \mdineq{\times} 640.
  
  \textbf{ScanNet200}~\cite{rozenberszki2022scannet200} extends ScanNetV2 to address a larger-vocabulary setting and provides 200 object categories for classification.
  The categories are divided into head, common, and tail groups, consisting of 66, 68, and 66 classes, respectively, based on the number of labeled surface points.
  Since smaller objects tend to have fewer surface points than larger ones, the average object size decreases from the head to the tail group (refer to \mdapp{A.1}). 
  We adopt the same train-test split as the one employed in ScanNet, and only objects of categories existing in both the training and the test sets are considered as in the official benchmark script.

  \mdsubsec{Evaluation Metric.}
  We adopt the mean average precision (mAP) to evaluate the detection accuracy.
  Specifically, we use mAP@0.25 and mAP@0.5 where the numbers indicate the 3D intersection over union (IoU) thresholds.
  To explain, the threshold is the minimum IoU to determine a positive match, which means mAP@0.5 is stricter on the evaluation of object location than mAP@0.25.
  
  \mdsubsec{Baselines.}
  We compare {\ourmethod} with ImVoxelNet~\cite{rukhovich2022imvoxelnet}, current state-of-the-art multi-view  image-based 3D detector, and VoteNet~\cite{qi2019votenet}, a seminal point cloud-based method.
  To reproduce ImVoxelNet, we use the code provided by the original authors~\cite{qi2019votenet}.
  As for VoteNet, we reproduce it with a common codebase~\cite{mmdet3d2020} and implement the same hyperparameters and model architecture as the original work~\cite{qi2019votenet}.

  \mdsubsec{Optimization.}
  The Adam optimizer~\cite{kingma2014adam} with a learning rate 0.0001 is used for training.
  The weight decay factor is 0.0001.
  Gradient clipping is also used and the max norm is set to 35.
  For ScanNetV2~\cite{dai2017scannet} and ARKitScenes~\cite{baruch2021arkitscenes}, the learning rate is reduced by ten times before the 9\textsuperscript{th} and 12\textsuperscript{th} epoch.
  For ScanNet200~\cite{rozenberszki2022scannet200}, the learning rate is reduced by ten times before the 9\textsuperscript{th} and 30\textsuperscript{th} epoch.
  The total numbers of epochs are 12, 12 and 30 for ScanNetV2, ARKitScenes, and ScanNet200, respectively.
  
  \mdsubsec{Voxel Volume.}
  The feature volume is 6.4 \mdineq{\times} 6.4 \mdineq{\times} 2.56 meters.
  The voxel size is 0.16 meters for ScanNetV2 and ARKitScenes, but 0.08 meters for ScanNet200 for detecting small objects.
  For ScanNetV2 and ScanNet200, we shift the volume origin to the coordinate origin, whereas for ARKitScenes, we relocate it to the center of the sampled poses.

  
  \begin{table}[ht!]
\small
\centering
\caption{
  \textbf{3D object detection results on ARKitScenes} are presented to demonstrate the practicability of {\ourmethod} in real-world scenarios.
  We also include the result (a-4) reported by \cite{baruch2021arkitscenes} for reference.
  The notations RGB and PC signify the use of color images and point clouds, respectively, during the inference phase.
  \vspace{5pt}
}
\begin{tabular}{l|cc|cc}
    \toprule
     & \multicolumn{2}{c|}{Input} & \multicolumn{2}{c}{Performance (mAP)} \\
    \cmidrule(lr){2-5}
    Method & RGB & PC & @0.25 & @0.5 \\
    \midrule
    \mdfootnotesize{(a-1)} {\ourmethod} (ours) & \checkmark & - & \textbf{60.2} & \textbf{43.4}\\
    \mdfootnotesize{(a-2)} 
    ImVoxelNet~\cite{rukhovich2022imvoxelnet} & \checkmark & - & 58.0 & 38.8\\
    \midrule
    \mdfootnotesize{(a-3)} VoteNet~\cite{qi2019votenet} & - & \checkmark & 53.3 & 38.5 \\
    \mdfootnotesize{(a-4)} VoteNet (from \cite{baruch2021arkitscenes}) & - & \checkmark & 35.8 & - \\
    \bottomrule
\end{tabular}
\label{table:arkit-result}
\end{table}

  \begin{table}[ht!]
\small
\centering
\caption{
  \textbf{3D object detection results on ScanNetV2}.
  In this benchmark, {\ourmethod} outperforms the SOTA multi-view image-based method, ImVoxelNet, by $12.5\%$ and $19.3\%$ in mAP@0.25 and mAP@0.5, respectively.
  In addition, the performance of {\ourmethod} with 50 views is not far from that of VoteNet, even though VoteNet leverages reconstructed meshes derived from typically more than 1000 viewpoints to obtain high-quality point clouds for its operations.
  \vspace{5pt}
}

\begin{tabular}{l|cc|cc}
    \toprule
     & \multicolumn{2}{c|}{Input} & \multicolumn{2}{c}{Performance (mAP)} \\
    \cmidrule(lr){2-5}
    Method & RGB & PC & @0.25 & @0.5 \\
    \midrule
    \mdfootnotesize{(s-1)} {\ourmethod} (ours) & \checkmark & - & \textbf{54.8} & \textbf{28.4} \\
    \mdfootnotesize{(s-2)} ImVoxelNet~\cite{rukhovich2022imvoxelnet} & \checkmark & - & 48.7 & 23.8 \\
    \midrule
    \mdfootnotesize{(s-3)} 3D-SIS~\cite{hou20193dsis} & - & \checkmark & 25.4 & 14.6 \\
    \mdfootnotesize{(s-4)} 3D-SIS~\cite{hou20193dsis} & \checkmark & \checkmark & 40.2 & 22.5 \\
    \mdfootnotesize{(s-5)} VoteNet~\cite{qi2019votenet} & - & \checkmark & 58.6 & 33.5 \\
    \bottomrule
\end{tabular}
\label{table:scannet-result}
\end{table}

  \begin{table}[ht!]
\small
\centering
\caption{
  \textbf{3D object detection results on ScanNet200} are presented to examine the robustness of {\ourmethod} for diverse classes.
  As in \cite{rozenberszki2022scannet200}, categories are divided into head, common and tail groups where the average object sizes decrease from the head to the tail group.
  \vspace{5pt}
}
\begin{tabular}{l|c|c|c|c|c}
    \toprule
     & & \multicolumn{4}{c}{Performance (mAP@0.25)} \\
    \cmidrule(lr){3-6}
     Method & Input & \mdfootnotesize{Total} & \mdfootnotesize{Head} & \mdfootnotesize{Comm} & \mdfootnotesize{Tail} \\
    \midrule
    \mdfootnotesize{(d-1)} {\ourmethod} (ours) & \mdfootnotesize{RGB} & \textbf{22.3} & \textbf{38.1} & \textbf{17.3} & \textbf{9.7}\\
    \mdfootnotesize{(d-2)} ImVoxelNet~\cite{rukhovich2022imvoxelnet} & \mdfootnotesize{RGB} & 19.0 & 34.1 & 14.0 & 7.7\\
    \midrule
    \mdfootnotesize{(d-3)} VoteNet~\cite{qi2019votenet} & \mdfootnotesize{PC} & 19.8 & 38.5 & 16.0 & 2.9 \\
    \bottomrule
\end{tabular}
\label{table:scannet200-result}
\end{table}


\begin{table*}[ht!]
\small
\centering
\caption{
  \textbf{3D object detection results for varying numbers of views} on ARKitScenes.
  {\ourmethod} shows great data efficiency by achieving results comparable to ImVoxelNet with 100 views while utilizing only 40 input views.
  \vspace{5pt}
}
\begin{tabular}{l|cc|c|c|c|c|c|c|c}
    \toprule
    & \multicolumn{2}{c|}{Input} & \multicolumn{7}{c}{Performance (mAP@0.25 / mAP@0.5)} \\
    \cmidrule(lr){2-10}
    Method & RGB & PC & 10 views & 20 views & 30 views & 40 views & 50 views & 75 views & 100 views\\
    \midrule
    \mdfootnotesize{(v-1)} {\ourmethod} (ours) & \checkmark & - & \textbf{39.0} / \textbf{21.9} & \textbf{53.1} / \textbf{34.3} & \textbf{57.1} / \textbf{39.2} & \textbf{59.5} / \textbf{42.7} & \textbf{60.2} / \textbf{43.4} & \textbf{61.8} / \textbf{45.0} & \textbf{62.4} / \textbf{45.7}\\
    \mdfootnotesize{(v-2)} ImVoxelNet~\cite{rukhovich2022imvoxelnet} & \checkmark & - & 36.2 / 19.6 & 50.5 / 30.6 & 54.6 / 35.2 & 57.4 / 37.9 & 58.0 / 38.8 & 58.8 / 40.5 & 59.7 / 42.0\\
    \midrule
    \mdfootnotesize{(v-3)} VoteNet~\cite{qi2019votenet} & - & \checkmark & 30.2 / 20.8 & 45.9 / 31.5 & 50.2 / 34.0 & 51.1 / 36.8 & 53.3 / 38.5 & 53.6 / 38.3 & 53.9 / 39.0\\
    \bottomrule
\end{tabular}
\label{table:num-view-result}
\end{table*}

  \begin{table}[ht!]
\small
\centering
\caption{
  \textbf{Inference time for different numbers of views}.
  The experiment is run on ARKitScenes with a single Nvidia 3090 GPU.
  The data loading time is ignored and the inference times are averaged over all test scenes.
  \vspace{5pt}
}
\begin{tabular}{l|c|c|c|c}
    \toprule
    & \multicolumn{4}{c}{Inference Time (ms)} \\
    \cmidrule(lr){2-5}
     & 20 & 40 & 50 & 100\\
    Method & views & views & views & views\\
    \midrule
    \mdfootnotesize{(t-1)} {\ourmethod} (ours) & 139.0 & 166.1 & 181.8 & 245.8\\
    \mdfootnotesize{(t-2)} ImVoxelNet~\cite{rukhovich2022imvoxelnet} & 113.1 & 140.0 & 155.9 & 219.7\\
    \bottomrule
\end{tabular}
\label{table:speed}
\end{table}

  \begin{table}[t!]
\small
\centering
\caption{
  \textbf{Effectiveness of the geometry shaping}.
  We conduct this study on ScanNetV2.
  Despite that (g-3) ImVoxelNet with MaGNet, a SOTA multi-view depth estimator, takes 5 times more input views, has 42\% larger model size and 15.7 times longer runtime, {\ourmethod} has better detection results.
  The model size and the inference time of {\ourmethod} on ScanNetV2 are 485.6 MB and 489.9 ms, respectively.
  \vspace{5pt}
}
\begin{tabular}{l|c|c|cc}
    \toprule
     & \multicolumn{2}{c|}{Relative} & \multicolumn{2}{c}{mAP} \\
    \cmidrule(lr){2-5}
    Method & Size & Runtime & @0.25 & @0.5 \\
    \midrule
    \mdfootnotesize{(g-1)} {\ourmethod} (ours) & 1.0 & 1.0 & \textbf{54.8} & \textbf{28.4} \\
    \midrule
    \mdfootnotesize{(g-2)} ImVoxelNet~\cite{rukhovich2022imvoxelnet} & 0.82 & 0.85 & 48.7 & 23.8 \\
    \mdfootnotesize{(g-3)} ~~w/ MaGNet~\cite{bae2022magnet} & 1.42 & 15.72 & 53.8 & 28.2 \\
    \mdfootnotesize{(g-4)} ~~w/ GT depth & 0.82 & 0.89 & 58.8 & 33.4 \\
    \bottomrule
\end{tabular}
\label{table:scannet-ablation-geo-shaping}
\end{table}

  \subsection{Results}
\label{subsec:results}

\begin{figure*}[ht!]
\begin{center}
  \includegraphics[width=\textwidth]{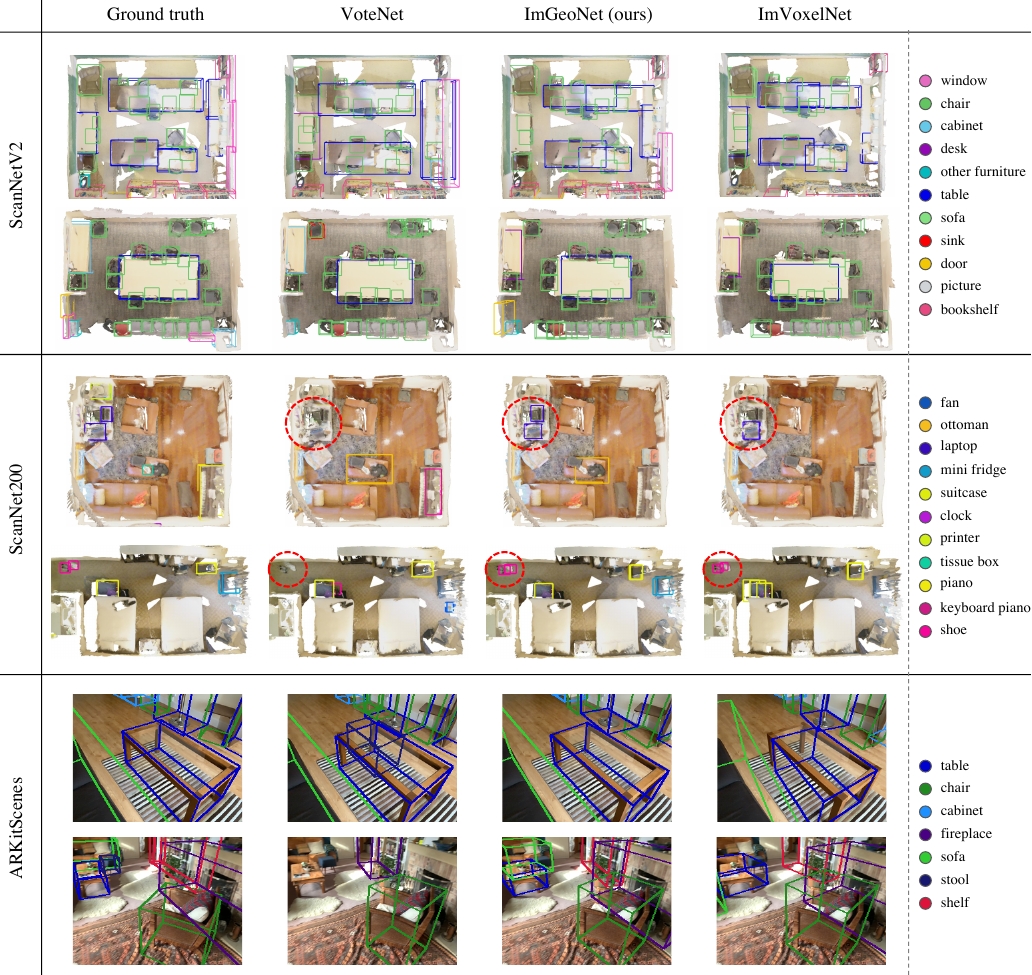}
\end{center}
\vspace{-5pt}
\caption{
  \textbf{Qualitative results of 3D object detection} on ScanNetV2, ScanNet200 and ARKitScenes.
  For ScanNet200, only objects in common or tail groups are presented.
  {\ourmethod} outperforms the multi-view image-based SOTA method, ImVoxelNet, on all three datasets.
  Compared to the seminal point cloud-based method, VoteNet, {\ourmethod} has superior results on small objects such as laptops and shoes (see the red circles in ScanNet200).
  Besides, in a more realistic mobile scenario (ARKitScenes), {\ourmethod} yields the most precise outcome.
}
\label{fig:vis}
\end{figure*}

  \mdsubsec{Realistic Mobile Capture.}
  First of all, we conduct an experiment on ARKitScenes which is captured by popular mobile devices. 
  The results are presented in \mdtbl{\ref{table:arkit-result}} and \mdfig{\ref{fig:vis}}.
  In the present scenario, wherein a test set comprising 50 evenly sampled views, {\ourmethod} (a-1) attains the best performance in mAP@0.25 and mAP@0.5.
  Through a comparison of {\ourmethod} (a-1) with the current state-of-the-art multi-view image-based method, ImVoxelnet (a-2), we demonstrate that the geometry-aware representation utilized in {\ourmethod} is effective in a practical mobile environment, where the depth maps used for training the geometry shaping network are not entirely accurate.
  Additionally, both image-based approaches (a-1 and a-2) exhibit superior performance compared to the seminal point cloud-based method, VoteNet (a-3).
  This highlights the practical preference for the use of images in real-world scenarios.

  
  \mdsubsec{Comparison on ScanNetV2.}
  Secondly, we compare our method with state-of-the-art methods on ScanNetV2, a well-known object detection benchmark for indoor scenes.
  The results are presented in \mdtbl{\ref{table:scannet-result}} and \mdfig{\ref{fig:vis}}.
  It can be observed from the results that {\ourmethod} (s-1) exhibits superior performance in comparison to ImVoxelNet (s-2), the current state-of-the-art multi-view image-based method.
  Specifically, {\ourmethod} outperforms ImVoxelNet by $12.5\%$ and $19.3\%$ in mAP@0.25 and mAP@0.5, respectively.
  The efficacy of our proposed geometry-aware representation is empirically validated through the significant improvement in mAP achieved by {\ourmethod}.
  On the other hand, {\ourmethod} (s-1) outperforms the point cloud-based baseline 3D-SIS (s-3) and its variant incorporating image data (s-4).
  Furthermore, {\ourmethod} successfully reduces the performance gap between the seminal point cloud-based method (s-5) and image-based methods to a significant extent.
  These are notable considering that point cloud-based methods rely on 3D reconstructed meshes that are obtained from a multitude of viewpoints (typically 1000+) to acquire high-quality point clouds for their operations, while {\ourmethod} only utilizes 50 views.

  \mdsubsec{Diverse Classes.}
  To inspect the capability of the proposed method for a diverse range of object classes, we conduct an analysis on ScanNet200, and the results are presented in \mdtbl{\ref{table:scannet200-result}} and \mdfig{\ref{fig:vis}}.
  In this scenario, {\ourmethod} (d-1) attains superior performance over ImVoxelNet (d-2) across all category groups, reaffirming the effectiveness of our geometry-aware volume representation in scenes containing objects of diverse classes.
  Furthermore, {\ourmethod} (d-1) exhibits superior performance over VoteNet (d-3), particularly for the common and tail category groups, which are characterized by smaller average object sizes in contrast to the head category group (refer to \mdapp{A.1}). 
  This achievement is remarkable, given that VoteNet utilizes vertices from high-quality 3D reconstructed meshes obtained from a great number of views (typically 1000+), while {\ourmethod} directly takes images from 50 viewpoints as input.
  We conjecture there are two main reasons for it:
  (1) The downsampling technique employed in the point cloud-based baseline has a tendency to exclude small instances~\cite{yang20203dssd}, thereby impeding the detection performance.
  (2) {\ourmethod} primarily relies on the geometry-aware representation induced by images, which effectively leverages the visual features extracted from a 2D backbone pre-trained on examples containing objects of various sizes.
  As a result, {\ourmethod} exhibits greater robustness for small objects.

  \mdsubsec{Number of Views.}
  In order to investigate the impact of varying numbers of views, we utilize different numbers of views in the construction of feature volume for image-based methods, as well as in the generation of point clouds for point cloud-based methods.
  We choose ARKitScenes as the dataset as its captured scenes are more representative of real-world scenarios.
  The results are presented in \mdtbl{\ref{table:num-view-result}}.
  First of all, we can observe that {\ourmethod} (v-1) with only 30 views outperforms VoteNet (v-3) with 100 views.
  Secondly, with only 40 views, the proposed geometry shaping enables {\ourmethod} (v-1) to achieve comparable performance to ImVoxelNet (v-2) with 100 views.
  The effectiveness of {\ourmethod} in utilizing data is particularly valuable in scenarios where only a limited number of views can be obtained.

  \mdsubsec{Model Speed.}
  Regarding inference time, we present a comparison between {\ourmethod} and ImVoxelNet on ARKitScenes in \mdtbl{\ref{table:speed}}.
  The experiment is run by a single Nvidia 3090 GPU and the data loading time is ignored.
  It can be observed that {\ourmethod} (t-1) is slightly slower than ImVoxelNet (t-2) when using the same number of views.
  However, as previously mentioned, the performance of {\ourmethod} (v-1) with only 40 views is comparable to that of ImVoxelNet (v-2) with 100 views.
  Notably, our method (t-1) achieves this level of performance with a significantly shorter inference time than ImVoxelNet (t-2) by 53.6 ms (a $24\%$ relative speed-up).
  This finding highlights the effectiveness of the proposed geometry-aware representation, as it demonstrates that {\ourmethod} can achieve a large performance improvement while only incurring a slight increase in running time.

  \mdsubsec{Effectiveness of Geometry Shaping Network.}
  To highlight the light overhead of the proposed Geometry Shaping Network (\mdineq{g(\cdot)} in \mdeq{\ref{eq:geo-shaping-grid}}), we extend ImVoxelNet with estimated depth (g-3) generated by MaGNet~\cite{bae2022magnet} (referred to as cascade baseline), the state-of-the-art multi-view depth estimator, and present the results on ScanNetV2 in \mdtbl{\ref{table:scannet-ablation-geo-shaping}}.
  Although both {\ourmethod} (g-1) and the cascade baseline (g-3) achieved similar levels of performance, the cascade baseline requires more resources.
  Specifically, MaGNet~\cite{bae2022magnet} requires four close vicinity views to produce a reliable depth map, resulting in five times more input views than {\ourmethod}. 
  Additionally, the cascade baseline has a total model size that is $42\%$ larger than {\ourmethod}.
  Furthermore, since MaGNet does not share features with ImVoxelNet, the inference time of the cascade baseline is 15.7 times longer than {\ourmethod}.
  
  On the other hand, ImVoxelNet with ground-truth depth (g-4) can be regarded as the upper limit.
  A sizable gap between {\ourmethod} (g-1) and the upper bound (g-4) can be observed, which indicates that there is room for improving Geometry Shaping Network in the future.
  Finally, we conduct a similar experiment on ARKitScenes, which shows the performance of the upper limit is 62.2 and 46.4 in mAP@0.25 and mAP@0.5, respectively.
  The performance gaps between the upper limit and {\ourmethod} in ARKitScenes (2.0/3.0 for mAP@0.25/@0.5) is smaller than those (4.0/5.0 for mAP@0.25/@0.5) in ScanNetV2.
  This observation serves to corroborate the effectiveness of the Geometry Shaping Network in real-world scenarios where the point clouds exhibit sparsity and noise.


\section{Conclusion}
  In this work, we have introduced {\ourmethod}, a 3D object detection framework that utilizes a geometry-aware voxel representation induced by multi-view images to model a 3D space.
  Since {\ourmethod} learns to predict geometry from multi-view images, a pre-trained 2D feature extractor can be leveraged and only images from multiple views are required during the inference phase.
  Through in-depth quantitative and qualitative experiments, we have demonstrated the effectiveness of our proposed geometry-aware representation by (1) achieving state-of-the-art results in image-based indoor 3D object detection, (2) showing great data efficiency by achieving great accuracy with fewer views, and (3) enabling image-based methods to attain superior detection accuracy than a seminal point cloud-based approach in practical scenarios with sparse and noisy point clouds or diverse object classes.
  
\section{Acknowledgements}
  This work is supported in part by Ministry of Science and Technology of Taiwan (NSTC 111-2634-F-002-022).
  We thank National Center for High-performance Computing (NCHC) for computational and storage resource.

{\small
\bibliographystyle{ieee_fullname}
\bibliography{reference}
}

\appendix


\section{Further Experiment}
\label{appendix:further-exp}
\subsection{Object Sizes in ScanNet200}
\label{appendix:scannet200-obj-size}
\begin{table}[ht!]
\small
\centering
\caption{
\textbf{Mean object sizes (m\textsuperscript{3}) across different groups in ScanNet200}, which are determined from the training data.
}

\vspace{5pt}



\begin{tabular}{cccc}
    \toprule
    & \multicolumn{3}{c}{ScanNet200} \\
    \cmidrule(lr){2-4}
    & Head & Common & Tail \\
    \midrule
    Object size & 0.92 & 0.14 & 0.03 \\
    \bottomrule
\end{tabular}
\label{appendix:table:scannet200-obj-size}
\end{table}
We report the mean object sizes of different category groups within ScanNet200~\cite{rozenberszki2022scannet200}.
The results presented in \mdtbl{\ref{appendix:table:scannet200-obj-size}} suggest that objects within the common and tail groups tend to have smaller sizes when compared to those within the head group.

\subsection{Visualization}
\label{appendix:vis}
\begin{figure*}[ht!]
\begin{center}
  \includegraphics[width=\textwidth]{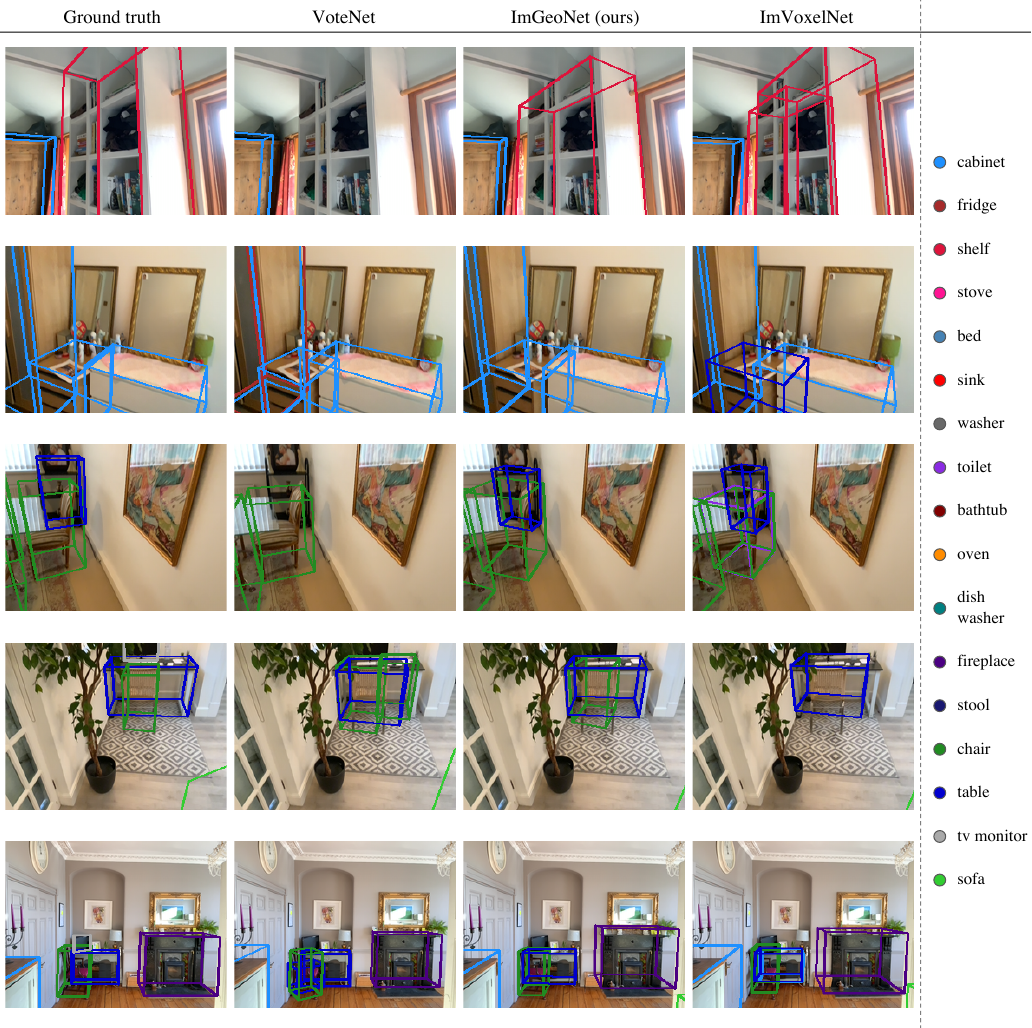}
\end{center}
\vspace{-7pt}
\caption{
  Qualitative results on ARKitScenes.
}
\label{appendix:fig:vis-more-arkit}
\end{figure*}

\begin{figure*}[ht!]
\begin{center}
  \includegraphics[width=\textwidth]{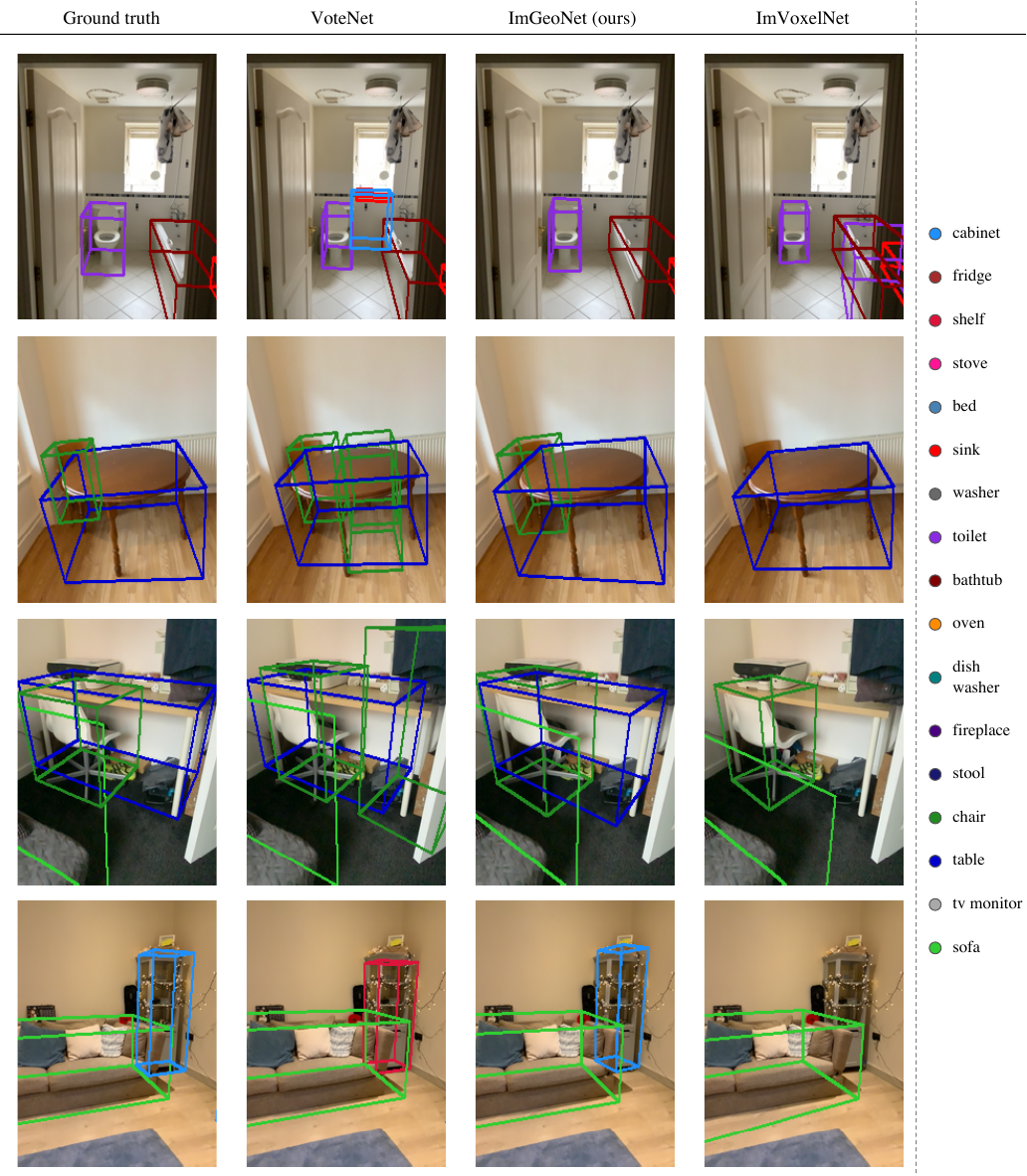}
\end{center}
\vspace{-7pt}
\caption{
  More qualitative results on ARKitScenes.
}
\label{appendix:fig:vis-more-arkit2}
\end{figure*}

\begin{figure*}[ht!]
\begin{center}
  \includegraphics[width=\textwidth]{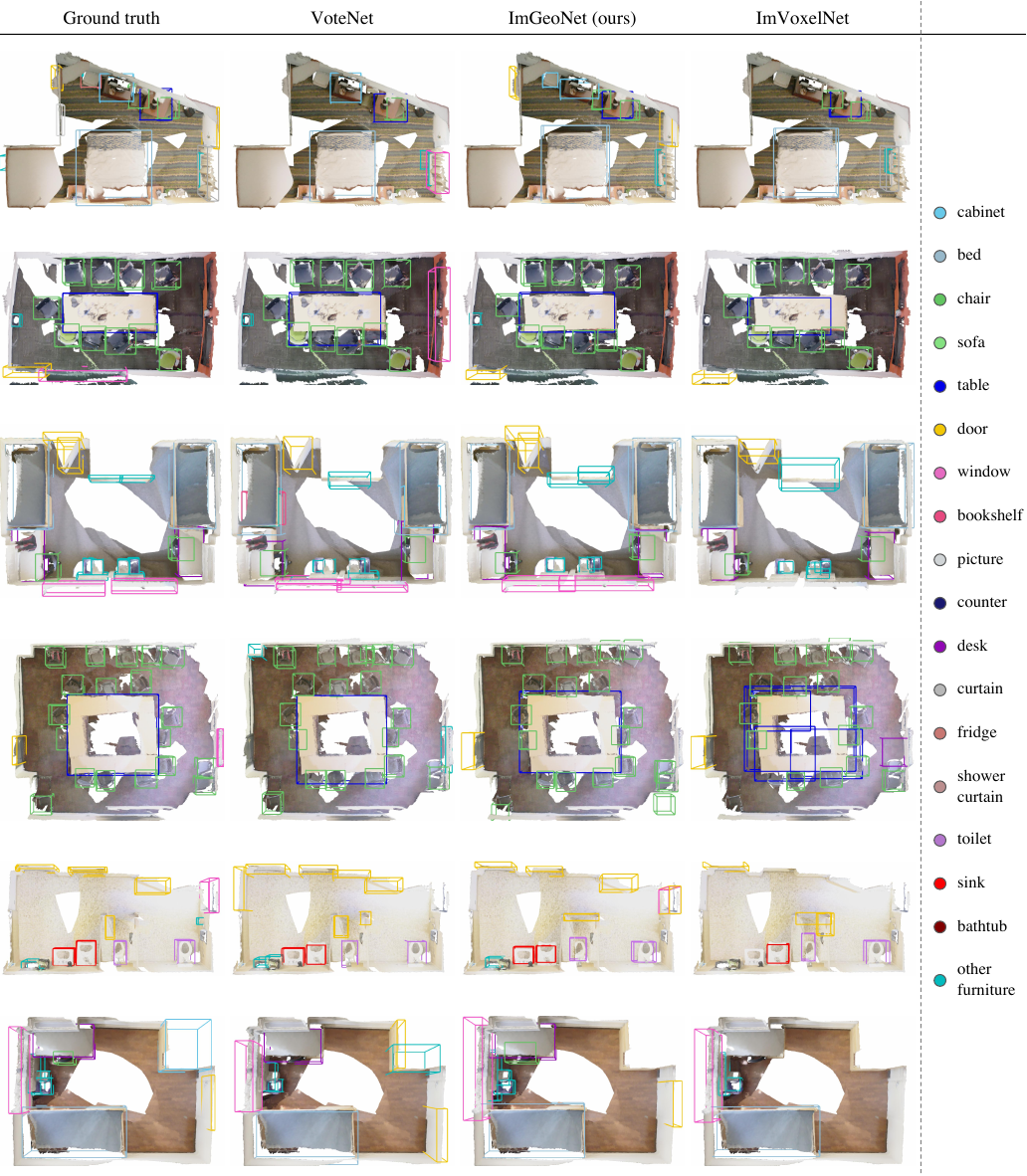}
\end{center}
\vspace{-7pt}
\caption{
  Qualitative results on ScanNetV2.
}
\label{appendix:fig:vis-more-scannet}
\end{figure*}

\begin{figure*}[ht!]
\begin{center}
  \includegraphics[width=\textwidth]{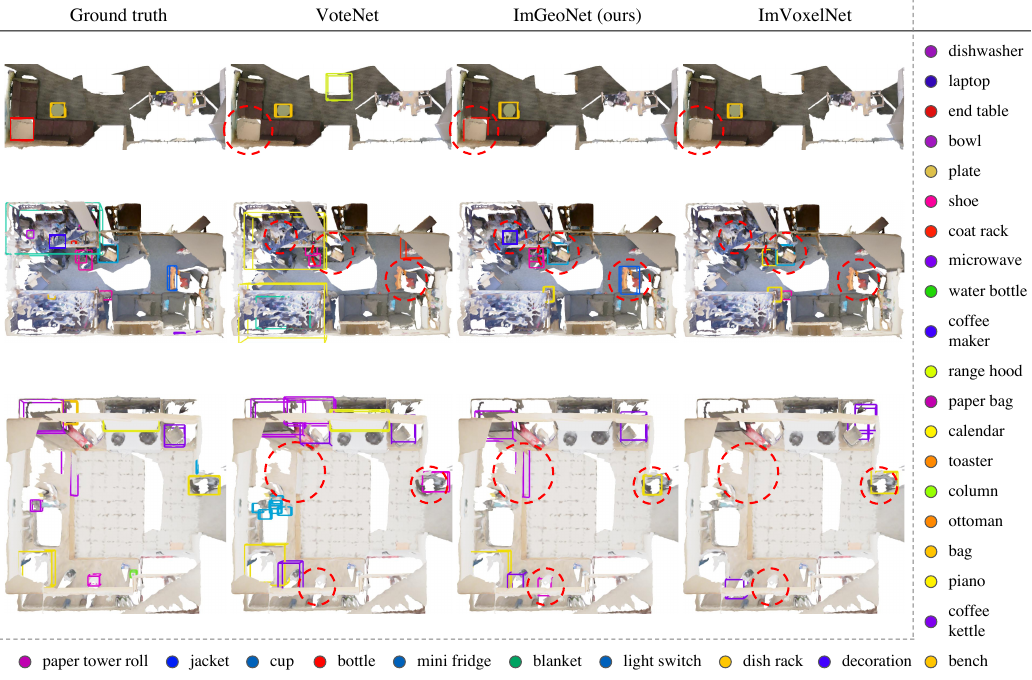}
\end{center}
\vspace{-7pt}
\caption{
  Qualitative results on ScanNet200.
  Only the common and tail category groups are presented.
}
\label{appendix:fig:vis-more-scannet200}
\end{figure*}

  This study presents the qualitative results of 3D object detection on three distinct datasets, namely ARKitScenes (as depicted in \mdfig{\ref{appendix:fig:vis-more-arkit}} and \mdfig{\ref{appendix:fig:vis-more-arkit2}}), ScanNetV2 (as illustrated in \mdfig{\ref{appendix:fig:vis-more-scannet}}), and ScanNet200 (as shown in \mdfig{\ref{appendix:fig:vis-more-scannet200}}).
  The effectiveness of our proposed geometry-aware representation can be verified by {\ourmethod}'s superior outcomes, in comparison to ImVoxelNet~\cite{rukhovich2022imvoxelnet}, across all datasets. Additionally, {\ourmethod} generates more precise results than the seminal point cloud-based approach, VoteNet~\cite{qi2019votenet}, in practical scenarios that involve sparse and noisy point clouds or diverse object classes.

\subsection{Per-class Evaluation}
  \label{appendix:per-cls-eval}

  We present the 3D object detection scores per class across different datasets.
  The outcomes obtained from ARKitScenes are provided in \mdtbl{\ref{appendix:table:exp-arkit}}.
  Similarly, the outcomes obtained from ScanNetV2 are presented in \mdtbl{\ref{appendix:table:exp-scannet}}
  Finally, the results for ScanNet200 are provided separately for three category groups: head, common, and tail.
  They are presented in \mdtbl{\ref{appendix:table:exp-scannet200-head}}, \mdtbl{\ref{appendix:table:exp-scannet200-common}}, and \mdtbl{\ref{appendix:table:exp-scannet200-tail}}, respectively.

\begin{table*}[ht!]
\small
\centering
\caption{
3D object detection scores per class on ARKitScenes.
}

\vspace{5pt}

\resizebox{1 \textwidth}{!}{
\setlength\tabcolsep{3pt} 
\begin{tabular}{l|ccccccccccccccccc|c}
    \toprule
    & cab & refrig & shelf & stove & bed & sink & washr & toil & bath & oven & dwashr & firepl & stool & chair & tabl & tvmnt & sofa & mAP@0.25 \\
    \midrule
    {\ourmethod} (ours) & 55.8 & 82.6 & 48.4 & 20.4 & 89.3 & 52.8 & 80.0 & 92.5 & 94.7 & 66.0 & 18.1 & 68.8 & 30.6 & 72.3 & 70.3 & 2.2 & 79.0 & \textbf{60.2} \\
    ImVoxelNet & 53.1 & 82.3 & 42.4 & 13.5 & 87.5 & 50.4 & 77.4 & 92.3 & 94.6 & 64.1 & 28.6 & 55.1 & 25.1 & 72.9 & 66.8 & 2.0 & 77.7 & 58.0 \\
    \midrule
    VoteNet & 62.1 & 79.6 & 26.0 & 4.9 & 92.9 & 43.5 & 56.5 & 91.9 & 94.1 & 37.4 & 6.4 & 67.8 & 18.6 & 57.4 & 62.8 & 21.6 & 83.2 & 53.3 \\
    \bottomrule
    \toprule
     & cab & refrig & shelf & stove & bed & sink & washr & toil & bath & oven & dwashr & firepl & stool & chair & tabl & tvmnt & sofa & mAP@0.5 \\
    \midrule
    {\ourmethod} (ours) & 31.8 & 72.5 & 21.7 & 3.9 & 83.3 & 19.9 & 71.2 & 84.8 & 91.0 & 44.4 & 15.9 & 23.1 & 13.3 & 49.3 & 45.1 & 0.1 & 67.2 & \textbf{43.4} \\
    ImVoxelNet & 25.2 & 68.8 & 11.3 & 3.9 & 75.9 & 10.9 & 61.2 & 82.5 & 89.6 & 38.3 & 24.7 & 13.4 & 11.0 & 42.3 & 36.0 & 0.0 & 64.2 & 38.8 \\
    \midrule
    VoteNet & 37.3 & 75.5 & 15.2 & 0.5 & 89.7 & 20.7 & 50.3 & 70.6 & 93.7 & 15.4 & 2.8 & 28.7 & 10.4 & 30.3 & 40.2 & 1.3 & 71.7 & 38.5 \\
    \bottomrule
\end{tabular}
}
\label{appendix:table:exp-arkit}
\end{table*}
\begin{table*}[ht!]
\small
\centering
\caption{
  3D object detection scores per class on ScanNetV2.
}

\vspace{5pt}

\resizebox{1 \textwidth}{!}{
\setlength\tabcolsep{3pt} 
\begin{tabular}{l|cccccccccccccccccc|c}
    \toprule
     & cab & bed & chair & sofa & tabl & door & wind & bkshf & pic & cntr & desk & curt & fridg & showr & toil & sink & bath & ofurn & mAP@0.25 \\
    \midrule
    {\ourmethod} (ours) & 38.7 & 86.5 & 76.6 & 75.7 & 59.3 & 42.0 & 28.1 & 59.2 & 4.3 & 42.8 & 71.5 & 36.9 & 51.8 & 44.1 & 95.2 & 58.0 & 79.6 & 36.8 & \textbf{54.8} \\
    ImVoxelNet & 34.4 & 81.9 & 74.6 & 71.9 & 55.1 & 32.0 & 16.9 & 38.2 & 2.0 & 37.0 & 62.2 & 26.5 & 54.1 & 39.0 & 90.7 & 52.1 & 74.9 & 32.8 & 48.7 \\
    \midrule
    VoteNet & 36.3 & 87.9 & 88.7 & 89.6 & 58.8 & 47.3 & 38.1 & 44.6 & 7.8 & 56.1 & 71.7 & 47.2 & 45.4 & 57.1 & 94.9 & 54.7 & 92.1 & 37.2 & 58.7 \\
    \bottomrule
    \toprule
     & cab & bed & chair & sofa & tabl & door & wind & bkshf & pic & cntr & desk & curt & fridg & showr & toil & sink & bath & ofurn & mAP@0.5 \\
    \midrule
    {\ourmethod} (ours) & 14.3 & 74.2 & 47.4 & 46.9 & 41.0 & 8.1 & 2.0 & 26.9 & 0.5 & 6.6 & 44.7 & 4.4 & 28.2 & 3.9 & 71.0 & 25.9 & 48.3 & 17.2 & \textbf{28.4} \\
    ImVoxelNet & 11.0 & 69.3 & 42.3 & 30.1 & 32.2 & 4.8 & 0.9 & 11.3 & 0.2 & 2.7 & 39.2 & 4.0 & 23.1 & 3.2 & 67.5 & 20.3 & 55.5 & 10.9 & 23.8 \\
    \midrule
    VoteNet & 8.1 & 76.1 & 67.2 & 68.8 & 42.4 & 15.3 & 6.4 & 28.0 & 1.3 & 9.5 & 37.5 & 11.6 & 27.8 & 10.0 & 86.5 & 16.8 & 78.9 & 11.7 & 33.5 \\
    \bottomrule
\end{tabular}
}
\label{appendix:table:exp-scannet}
\end{table*}

\begin{table*}[ht!]
\small
\centering
\caption{
  3D object detection scores per class on ScanNet200's \textbf{head} group (IoU threshold = 0.25).
}

\vspace{5pt}

\resizebox{1 \textwidth}{!}{
\setlength\tabcolsep{3pt} 
\begin{tabular}{l|cccccccc}
	\toprule
	& wall & chair & floor & table & door & couch & cabinet & shelf\\
	\midrule
	{\ourmethod} (ours) & 28.4 & 65.9 & 41.7 & 42.0 & 42.0 & 67.6 & 22.5 & 25.9\\
	ImVoxelNet & 18.9 & 64.0 & 43.5 & 39.0 & 31.1 & 69.4 & 21.8 & 19.9\\
	\midrule
	VoteNet & 46.5 & 79.0 & 96.6 & 50.3 & 53.9 & 89.5 & 34.1 & 35.7\\
	\bottomrule
	\toprule
	& desk & office chair & bed & pillow & sink & picture & window & toilet\\
	\midrule
	{\ourmethod} (ours) & 70.5 & 24.0 & 96.0 & 61.5 & 61.9 & 10.9 & 25.2 & 92.9\\
	ImVoxelNet & 64.4 & 20.7 & 91.5 & 45.4 & 56.2 & 4.2 & 12.8 & 93.0\\
	\midrule
	VoteNet & 71.9 & 26.8 & 93.2 & 27.2 & 47.2 & 6.1 & 34.0 & 96.7\\
	\bottomrule
	\toprule
	& bookshelf & monitor & curtain & book & armchair & coffee table & box & refrigerator\\
	\midrule
	{\ourmethod} (ours) & 52.2 & 68.0 & 30.6 & 33.7 & 57.7 & 74.3 & 24.9 & 49.5\\
	ImVoxelNet & 53.6 & 58.1 & 20.8 & 21.9 & 54.8 & 69.5 & 18.7 & 52.7\\
	\midrule
	VoteNet & 34.2 & 56.3 & 35.5 & 6.8 & 51.1 & 78.4 & 5.2 & 63.5\\
	\bottomrule
	\toprule
	& lamp & kitchen cabinet & towel & clothes & tv & nightstand & counter & dresser\\
	\midrule
	{\ourmethod} (ours) & 41.2 & 24.4 & 19.5 & 11.7 & 57.5 & 76.5 & 24.0 & 41.5\\
	ImVoxelNet & 35.3 & 23.3 & 11.8 & 5.5 & 43.3 & 71.0 & 9.2 & 44.3\\
	\midrule
	VoteNet & 34.1 & 33.8 & 15.1 & 9.0 & 49.2 & 71.1 & 24.8 & 22.5\\
	\bottomrule
	\toprule
	& stool & plant & ceiling & bathtub & backpack & tv stand & whiteboard & shower curtain\\
	\midrule
	{\ourmethod} (ours) & 15.2 & 45.5 & 0.0 & 71.2 & 62.9 & 52.3 & 27.9 & 38.3\\
	ImVoxelNet & 13.6 & 29.3 & 0.0 & 77.0 & 62.2 & 48.1 & 14.1 & 32.0\\
	\midrule
	VoteNet & 21.9 & 22.7 & 36.1 & 86.8 & 32.6 & 35.0 & 24.2 & 63.2\\
	\bottomrule
	\toprule
	& trash can & closet & stairs & stove & board & washing machine & mirror & copier\\
	\midrule
	{\ourmethod} (ours) & 59.4 & 10.1 & 43.3 & 70.6 & 5.1 & 48.8 & 12.4 & 79.8\\
	ImVoxelNet & 53.2 & 7.8 & 32.8 & 64.4 & 1.8 & 56.7 & 6.3 & 72.2\\
	\midrule
	VoteNet & 41.7 & 3.9 & 15.1 & 75.7 & 4.2 & 48.5 & 14.3 & 91.7\\
	\bottomrule
	\toprule
	& sofa chair & file cabinet & shower & blinds & blackboard & radiator & recycling bin & wardrobe\\
	\midrule
	{\ourmethod} (ours) & 2.6 & 39.7 & 6.3 & 0.0 & 45.4 & 33.0 & 60.3 & 12.6\\
	ImVoxelNet & 5.7 & 33.8 & 16.8 & 0.0 & 44.4 & 25.8 & 51.4 & 9.4\\
	\midrule
	VoteNet & 10.1 & 16.1 & 26.9 & 1.4 & 51.1 & 37.8 & 7.3 & 38.4\\
	\bottomrule
	\toprule
	& clothes dryer & bathroom stall & shower wall & kitchen counter & doorframe & mailbox & object & bathroom vanity\\
	\midrule
	{\ourmethod} (ours) & 0.0 & 47.0 & 56.8 & 31.8 & 18.2 & 7.0 & 1.2 & 44.3\\
	ImVoxelNet & 0.0 & 48.2 & 58.1 & 24.6 & 16.6 & 4.2 & 0.5 & 41.7\\
	\midrule
	VoteNet & 0.4 & 43.5 & 60.1 & 20.6 & 30.1 & 12.5 & 0.3 & 57.4\\
	\bottomrule
	\toprule
	& closet wall & stair rail & \textbf{mAP}\\
	\midrule
	{\ourmethod} (ours) & 0.1 & 1.1 & \textbf{38.1}\\
	ImVoxelNet & 4.5 & 1.9 & 34.1\\
	\midrule
	VoteNet & 17.2 & 10.9 & 38.5\\
	\bottomrule
\end{tabular}

}
\label{appendix:table:exp-scannet200-head}
\end{table*}

\begin{table*}[ht!]
\small
\centering
\caption{
  3D object detection scores per class on ScanNet200's \textbf{common} group (IoU threshold = 0.25).
}

\vspace{5pt}

\resizebox{1 \textwidth}{!}{
\setlength\tabcolsep{3pt} 

\begin{tabular}{l|ccccccc}
	\toprule
	& cushion & end table & dining table & keyboard & bag & toilet paper & printer\\
	\midrule
	{\ourmethod} (ours) & 0.0 & 19.9 & 7.8 & 18.4 & 12.8 & 17.7 & 41.7\\
	ImVoxelNet & 0.0 & 17.8 & 3.3 & 5.5 & 10.4 & 12.1 & 29.4\\
	\midrule
	VoteNet & 0.1 & 17.2 & 5.7 & 0.4 & 2.7 & 2.1 & 29.6\\
	\bottomrule
	\toprule
	& blanket & microwave & shoe & computer tower & bottle & bin & ottoman\\
	\midrule
	{\ourmethod} (ours) & 19.3 & 51.5 & 36.3 & 21.3 & 1.1 & 0.5 & 51.2\\
	ImVoxelNet & 6.5 & 31.5 & 27.0 & 17.8 & 2.3 & 0.2 & 54.6\\
	\midrule
	VoteNet & 0.7 & 23.4 & 4.6 & 45.0 & 0.3 & 1.7 & 56.6\\
	\bottomrule
	\toprule
	& bench & basket & fan & laptop & person & paper towel dispenser & oven\\
	\midrule
	{\ourmethod} (ours) & 8.5 & 5.3 & 12.3 & 76.0 & 8.7 & 72.2 & 17.7\\
	ImVoxelNet & 8.0 & 13.3 & 7.4 & 62.4 & 9.9 & 81.6 & 6.6\\
	\midrule
	VoteNet & 21.2 & 0.1 & 3.0 & 0.4 & 14.5 & 63.3 & 0.8\\
	\bottomrule
	\toprule
	& rack & piano & suitcase & rail & container & telephone & stand\\
	\midrule
	{\ourmethod} (ours) & 0.0 & 30.6 & 48.2 & 0.4 & 2.1 & 24.4 & 0.0\\
	ImVoxelNet & 0.0 & 13.8 & 43.3 & 0.1 & 1.3 & 21.8 & 0.0\\
	\midrule
	VoteNet & 0.0 & 33.9 & 49.2 & 2.0 & 0.5 & 1.5 & 0.0\\
	\bottomrule
	\toprule
	& light & laundry basket & pipe & seat & column & ladder & jacket\\
	\midrule
	{\ourmethod} (ours) & 0.0 & 3.7 & 0.1 & 0.0 & 0.0 & 5.0 & 17.7\\
	ImVoxelNet & 0.0 & 4.8 & 0.0 & 0.0 & 0.0 & 14.0 & 14.6\\
	\midrule
	VoteNet & 1.7 & 1.1 & 0.0 & 1.4 & 30.9 & 1.1 & 0.2\\
	\bottomrule
	\toprule
	& storage bin & coffee maker & dishwasher & machine & mat & windowsill & bulletin board\\
	\midrule
	{\ourmethod} (ours) & 9.4 & 56.8 & 48.1 & 0.0 & 19.6 & 3.9 & 19.9\\
	ImVoxelNet & 5.5 & 26.5 & 27.9 & 0.0 & 9.8 & 0.3 & 11.8\\
	\midrule
	VoteNet & 0.9 & 12.9 & 4.8 & 0.4 & 0.0 & 7.5 & 2.5\\
	\bottomrule
	\toprule
	& fireplace & mini fridge & water cooler & shower door & pillar & ledge & furniture\\
	\midrule
	{\ourmethod} (ours) & 3.9 & 43.5 & 0.0 & 7.0 & 0.0 & 1.2 & 0.0\\
	ImVoxelNet & 9.6 & 36.1 & 1.7 & 18.3 & 0.0 & 1.0 & 0.0\\
	\midrule
	VoteNet & 0.3 & 22.8 & 2.3 & 20.4 & 5.7 & 0.0 & 0.1\\
	\bottomrule
	\toprule
	& cart & decoration & closet door & vacuum cleaner & dish rack & range hood & projector screen\\
	\midrule
	{\ourmethod} (ours) & 23.6 & 0.7 & 2.9 & 53.4 & 96.1 & 20.7 & 0.0\\
	ImVoxelNet & 18.4 & 0.0 & 0.6 & 16.9 & 84.2 & 14.0 & 0.0\\
	\midrule
	VoteNet & 31.2 & 20.0 & 0.3 & 19.1 & 60.8 & 4.0 & 6.0\\
	\bottomrule
	\toprule
	& divider & bathroom counter & laundry hamper & bathroom stall door & ceiling light & trash bin & bathroom cabinet\\
	\midrule
	{\ourmethod} (ours) & 19.5 & 0.0 & 8.9 & 1.2 & 0.0 & 18.9 & 35.8\\
	ImVoxelNet & 10.0 & 0.8 & 7.9 & 0.0 & 0.0 & 33.2 & 37.8\\
	\midrule
	VoteNet & 33.9 & 1.1 & 15.9 & 64.1 & 3.3 & 55.0 & 65.9\\
	\bottomrule
	\toprule
	& potted plant & mattress & \textbf{mAP}\\
	\midrule
	{\ourmethod} (ours) & 0.0 & 0.0 & \textbf{17.3}\\
	ImVoxelNet & 10.0 & 0.0 & 14.0\\
	\midrule
	VoteNet & 100.0 & 60.4 & 16.0\\
	\bottomrule
\end{tabular}

}
\label{appendix:table:exp-scannet200-common}
\end{table*}

\begin{table*}[ht!]
\small
\centering
\caption{
  3D object detection scores per class on ScanNet200's \textbf{tail} group (IoU threshold = 0.25).
}

\vspace{5pt}

\resizebox{1 \textwidth}{!}{
\setlength\tabcolsep{3pt} 
\begin{tabular}{l|ccccccc}
	\toprule
	& paper & plate & soap dispenser & bucket & clock & guitar & toilet paper holder\\
	\midrule
	{\ourmethod} (ours) & 3.0 & 0.0 & 15.8 & 9.4 & 0.3 & 74.6 & 0.0\\
	ImVoxelNet & 0.5 & 0.0 & 20.3 & 8.3 & 0.7 & 98.3 & 0.0\\
	\midrule
	VoteNet & 0.0 & 0.0 & 0.1 & 2.4 & 0.1 & 0.0 & 0.0\\
	\bottomrule
	\toprule
	& speaker & cup & paper towel roll & bar & toaster & ironing board & soap dish\\
	\midrule
	{\ourmethod} (ours) & 0.0 & 0.2 & 26.1 & 1.7 & 0.0 & 1.7 & 1.2\\
	ImVoxelNet & 0.0 & 0.2 & 23.0 & 1.8 & 0.0 & 0.0 & 0.0\\
	\midrule
	VoteNet & 0.0 & 0.0 & 0.3 & 0.2 & 0.0 & 0.1 & 0.0\\
	\bottomrule
	\toprule
	& toilet paper dispenser & fire extinguisher & ball & hat & shower curtain rod & paper cutter & tray\\
	\midrule
	{\ourmethod} (ours) & 47.3 & 0.0 & 25.1 & 0.0 & 0.0 & 59.9 & 1.0\\
	ImVoxelNet & 20.5 & 0.0 & 54.7 & 0.0 & 0.0 & 51.7 & 0.9\\
	\midrule
	VoteNet & 11.3 & 0.0 & 2.6 & 0.0 & 0.0 & 0.0 & 0.0\\
	\bottomrule
	\toprule
	& toaster oven & mouse & toilet seat cover dispenser & scale & tissue box & light switch & crate\\
	\midrule
	{\ourmethod} (ours) & 1.0 & 0.0 & 6.4 & 90.7 & 7.7 & 0.0 & 5.0\\
	ImVoxelNet & 17.3 & 0.0 & 0.0 & 60.6 & 12.2 & 0.0 & 10.0\\
	\midrule
	VoteNet & 24.4 & 0.0 & 6.2 & 0.0 & 0.0 & 0.0 & 0.3\\
	\bottomrule
	\toprule
	& power outlet & sign & projector & plunger & stuffed animal & headphones & broom\\
	\midrule
	{\ourmethod} (ours) & 0.0 & 12.4 & 0.0 & 0.0 & 0.0 & 0.0 & 0.0\\
	ImVoxelNet & 0.0 & 8.4 & 0.0 & 0.0 & 0.0 & 0.0 & 0.0\\
	\midrule
	VoteNet & 0.0 & 0.0 & 0.0 & 0.0 & 0.0 & 0.0 & 0.0\\
	\bottomrule
	\toprule
	& dustpan & hair dryer & water bottle & handicap bar & vent & shower floor & water pitcher\\
	\midrule
	{\ourmethod} (ours) & 0.0 & 0.0 & 0.0 & 0.0 & 0.0 & 50.0 & 0.0\\
	ImVoxelNet & 0.0 & 0.0 & 0.0 & 0.0 & 0.0 & 0.0 & 0.0\\
	\midrule
	VoteNet & 0.0 & 0.0 & 0.0 & 0.0 & 0.0 & 100.0 & 0.0\\
	\bottomrule
	\toprule
	& bowl & paper bag & laundry detergent & dumbbell & tube & closet rod & coffee kettle\\
	\midrule
	{\ourmethod} (ours) & 8.6 & 0.2 & 0.0 & 50.0 & 0.1 & 0.0 & 0.0\\
	ImVoxelNet & 2.9 & 0.0 & 0.0 & 30.0 & 0.1 & 0.0 & 0.0\\
	\midrule
	VoteNet & 0.0 & 0.7 & 0.0 & 0.0 & 0.0 & 0.0 & 0.0\\
	\bottomrule
	\toprule
	& shower head & keyboard piano & case of water bottles & coat rack & folded chair & fire alarm & power strip\\
	\midrule
	{\ourmethod} (ours) & 0.0 & 60.0 & 3.0 & 0.0 & 0.0 & 0.0 & 0.0\\
	ImVoxelNet & 0.0 & 13.2 & 0.0 & 0.0 & 0.0 & 0.0 & 0.0\\
	\midrule
	VoteNet & 0.0 & 0.3 & 4.8 & 11.1 & 1.0 & 0.0 & 0.0\\
	\bottomrule
	\toprule
	& calendar & poster & \textbf{mAP}\\
	\midrule
	{\ourmethod} (ours) & 0.0 & 0.0 & \textbf{9.7}\\
	ImVoxelNet & 6.7 & 2.9 & 7.7\\
	\midrule
	VoteNet & 0.0 & 0.0 & 2.9\\
	\bottomrule
\end{tabular}

}
\label{appendix:table:exp-scannet200-tail}
\end{table*}


\end{document}